\documentclass[letterpaper, 10 pt, conference]{ieeeconf}

\IEEEoverridecommandlockouts 
\usepackage{graphicx}
\usepackage{epsfig} 
\usepackage{amsmath} 

\usepackage{amssymb}  
\usepackage{amsthm}
\usepackage{wasysym}
\usepackage{mathrsfs}  
\usepackage{bbm}
\usepackage{color}
\usepackage[ruled,vlined,linesnumbered]{algorithm2e}
\usepackage[colorlinks=true,linkcolor=black,anchorcolor=black,citecolor=black,filecolor=black,menucolor=black,runcolor=black,urlcolor=black]{hyperref}
\usepackage{etoolbox}
\usepackage[table,xcdraw,dvipsnames]{xcolor}
\usepackage{multirow}
\usepackage{mathtools}

\usepackage{caption}
\usepackage{subcaption}

\usepackage{outlines}
\let\labelindent\relax
\usepackage{enumitem}
\setenumerate[1]{label=\arabic*.}
\setenumerate[2]{label=\alph*.}

\usepackage{algorithm2e}
\RestyleAlgo{ruled}

\newtheorem{defn}{Definition}
\newtheorem{rem}[defn]{Remark}

\newtheorem{prop}[defn]{Proposition}

\providecommand{\R}{\ensuremath \mathbb{R}}
\providecommand{\N}{\ensuremath \mathbb{N}}

\newcommand{\regtext}[1]{\mathrm{\textnormal{#1}}}

\newcommand{\ts}[1]{\textsuperscript{#1}}
\newcommand{\lbl}[1]{_{\regtext{#1}}}
\newcommand{\ith}{{$i$\ts{th}} } %
\newcommand{\jth}{{$j$\ts{th}} }

\newcommand{\card}[1]{\left\vert#1\right\vert}

\newcommand{\norm}[1]{\left\Vert#1\right\Vert}
\newcommand{\abs}[1]{\left\vert#1\right\vert}
\newcommand{\floor}[1]{\left\lfloor#1\right\rfloor}

\newcommand{\diag}{\regtext{diag}}

\newcommand{\union}{\bigcup}

\newcommand{\trans}{^\top}

\newcommand{\set}[1]{\mathcal{#1}}
\newcommand{\setfn}[1]{\mathscr{#1}}

\newcommand{\normaldist}{\mathscr{N}}
\newcommand{\vep}{\varepsilon}

\newcommand{\goal}{{\regtext{g}}}
\newcommand{\ego}{{\regtext{e}}}
\newcommand{\nom}{{\regtext{nom}}}
\newcommand{\obs}{{\regtext{obs}}}

\newcommand{\zeros}{{0}}
\newcommand{\ones}{{1}}
\newcommand{\eye}{{I}}

\newcommand{\idx}[1]{{[#1]}} 
\newcommand{\state}{x}
\newcommand{\mode}{y}
\newcommand{\ctrl}{u}
\newcommand{\noise}{w}
\newcommand{\position}{p}

\newcommand{\dyn}{f}

\newcommand{\horizon}{{t\lbl{f}}}
\newcommand{\consensushorizon}{{t\lbl{c}}}
\newcommand{\horizondisc}{{k\lbl{f}}}
\newcommand{\consensushorizondisc}{{k\lbl{c}}}

\newcommand{\dt}{{\Delta_t}}
\newcommand{\cont}[1]{\overline{#1}}

\newcommand{\zonotope}{\set{Z}}
\newcommand{\contzono}{\cont{\zonotope}}
\newcommand{\poszono}{\set{P}}

\newcommand{\reachset}{\set{R}}
\newcommand{\contreachset}{\cont{\reachset}}

\newcommand{\noiseset}{\set{W}}

\newcommand{\linesegment}{\set{L}}

\newcommand{\statespace}{\set{X}}
\newcommand{\ctrlspace}{\set{U}}
\newcommand{\varctrlspace}{\set{V}}

\newcommand{\modes}{\set{Y}}
\newcommand{\timeinterval}{\set{T}}

\newcommand{\statehistory}{X_0}

\newcommand{\ctrlfuture}{U\lbl{f}}

\newcommand{\costfunc}{J}

\newcommand{\posproj}{\regtext{pos}}

\newcommand{\nstate}{{n_\state}}
\newcommand{\nctrl}{{n_\ctrl}}

\newcommand{\nmodes}{{n_\mode}}
\newcommand{\nagents}{{m}}
\newcommand{\nego}{{n_{\ego}}}
\newcommand{\ngen}{{n_G}}

\newcommand{\nhistory}{{h}}

\newcommand{\modeprobability}{{\gamma}}

\newcommand{\zonofnplain}{{\setfn{Z}\!}}
\newcommand{\zonofn}[1]{{\zonofnplain\!\left(#1\right)}}

\newcommand{\pred}{\setfn{P}}
\newcommand{\param}{\theta}
\newcommand{\mean}{\mu}
\newcommand{\std}{\Sigma}

\newcommand{\variation}{\delta}
\newcommand{\expansion}[1]{\hat{#1}}

\usepackage[
    style=ieee,
    doi=false,
    isbn=false,
    url=false,
    eprint=false,
    backend=biber,
    natbib=true
    ]{biblatex}

\bibliography{references}

\title{\LARGE \bf
ZAPP!
Zonotope Agreement of Prediction and Planning for Continuous-Time Collision Avoidance with Discrete-Time Dynamics
}

\author{Luca Paparusso\ts{1},
Shreyas Kousik\ts{2},
Edward Schmerling\ts{3},
Francesco Braghin\ts{1},
Marco Pavone\ts{3,4}
\thanks{\ts{1}Dept. of Mechanical Engineering, Politecnico di Milano, Milan, Italy.}
\thanks{\ts{2}School of Mechanical Engineering, Georgia Tech, Atlanta, GA, USA.}
\thanks{\ts{3}NVIDIA Autonomous Vehicle Research Group, Santa Clara, CA, USA.}
\thanks{\ts{4}Aeronautics and Astronautics, Stanford University, Stanford, CA, USA.}
\thanks{Toyota Research Institute provided funds to support this work.
Corresponding author: Luca Paparusso, \texttt{luca.paparusso@polimi.it}.}
}

\begin{document}

\maketitle
\thispagestyle{empty}
\pagestyle{empty}

\begin{abstract}
The past few years have seen immense progress on two fronts that are critical to safe, widespread mobile robot deployment: predicting uncertain motion of multiple agents, and planning robot motion under uncertainty.
However, the numerical methods required on each front have resulted in a mismatch of representation for prediction and planning.
In prediction, numerical tractability is usually achieved by coarsely discretizing time, and by representing multimodal multi-agent interactions as distributions with infinite support.
On the other hand, safe planning typically requires very fine time discretization, paired with distributions with compact support, to reduce conservativeness and ensure numerical tractability.
The result is, when existing predictors are coupled with planning and control, one may often find unsafe motion plans.
This paper proposes ZAPP (Zonotope Agreement of Prediction and Planning) to resolve the representation mismatch.
ZAPP unites a prediction-friendly coarse time discretization and a planning-friendly zonotope uncertainty representation; the method also enables differentiating through a zonotope collision check, allowing one to integrate prediction and planning within a gradient-based optimization framework.
Numerical examples show how ZAPP can produce safer trajectories compared to baselines in interactive scenes.
\end{abstract}

\section{Introduction}

The widespread deployment of autonomous mobile robots near and around people is steadily increasing.
To ensure safety (i.e., collision avoidance), such robots require predictive models of other agents' uncertain motion \cite{rudenko2020human,leon2021review}, plus motion planning methods that can quickly generate collision-avoiding trajectories \cite{mohanan2018survey,gonzalez2015review}.
However, there is often a mismatch in the numerical representation required to implement predictors and to implement planners and controllers.
In particular, predictors typically leverage a coarse time discretization and represent uncertain agent motion via Gaussian mixtures or other unbounded distributions \cite{salzmann2020trajectron,IvanovicElhafsiEtAl2020,jacobs2017real,shi2022motion,kamenev2022predictionnet,gilles2022gohome}.
In contrast, planning typically requires a fine time discretization to avoid excessive conservativeness and accurately represent dynamics \cite{claussmann2019review,sanchez2002delaying,elbanhawi2014sampling}.
Furthermore, since safety assurances are nominally incompatible with unbounded representations of agent motion, planning methods typically consider bounded disturbances (e.g., \cite{danielson2020robust,kousik2020bridging,chen2021fastrack,pek2020fail}).
Thus, it remains unclear how best to bridge the representation gap between prediction and motion planning for mobile robots in multi-agent scenes.

We propose a Zonotope Agreement of Prediction and Planning (ZAPP) to address the representation gap between prediction and planning.
ZAPP seeks to preserve as many of the learned properties of multi-agent interaction from a predictor while producing a computationally-tractable, planning-compatible representation.
We reformulate a generalized notion of prediction using zonotopes, a convex, symmetric polytope representation amenable to robust planning and control, which lets us represent uncertainty and continuous-time motion.
ZAPP combines these components into a trajectory optimization framework inspired by recent successes in gradient-based safe motion planning \cite{schaefer2021leveraging,SelimAlanwarEtAl2022,holmes2020reachable,kousik2019safe}.
We also note that our approach is predictor-agnostic, meaning it can be paired with a variety of existing discrete-time predictors.

\begin{figure}[t]
    \centering
    \includegraphics[width=0.90\columnwidth]{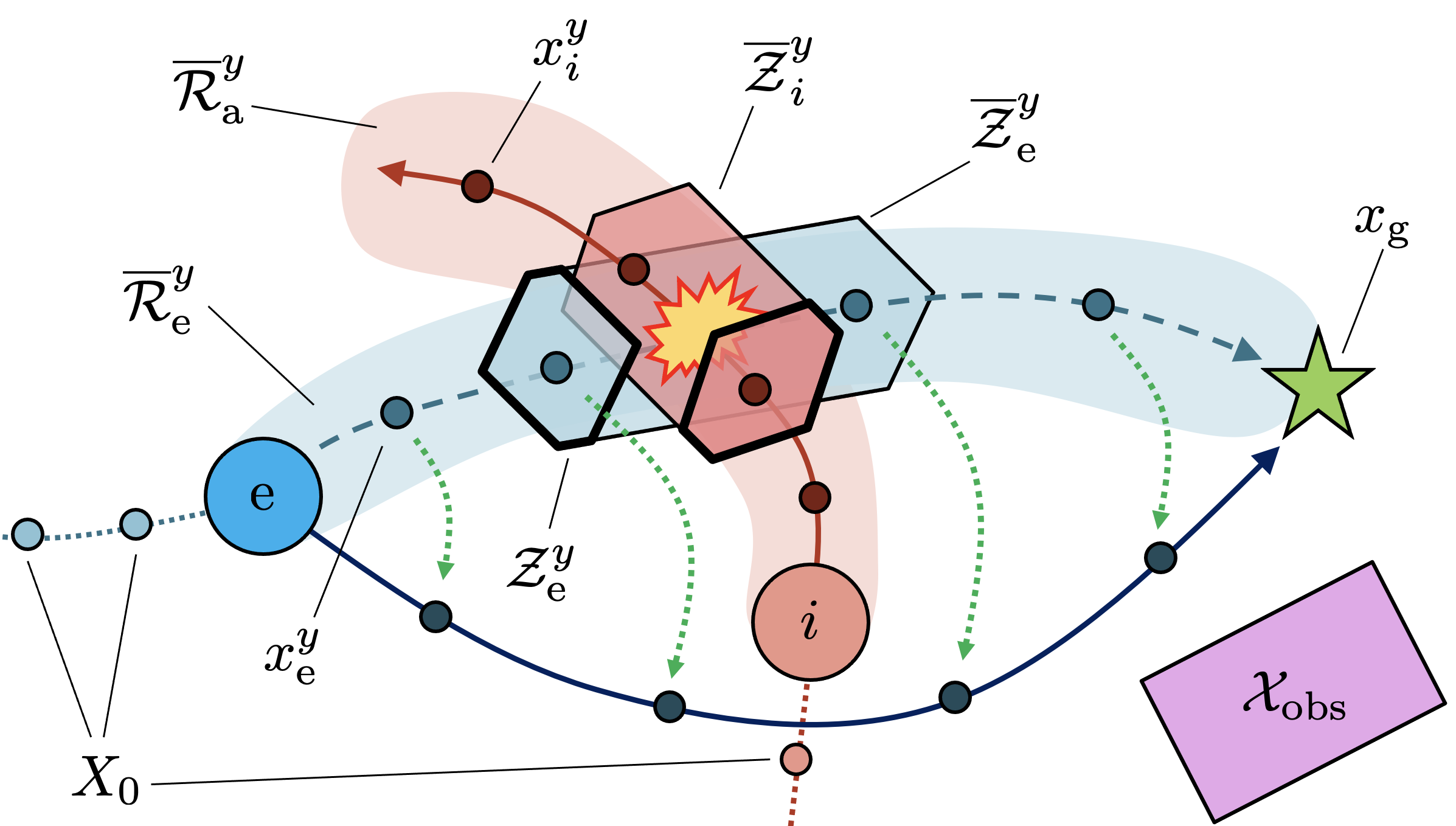}
    \caption{\small
    Overview of approach and notation.
    Existing interaction prediction approaches typically use coarse time discretizations to enable long-term predictions, but this can lead to missed collision detections in motion planning.
    We use zonotopes (blue and red polygons) to represent uncertain, continuous-time reachable sets (light blue and red tubes) for agents in interactive scenes (ego agent in blue, other agent $i$ in red).
    Note that approximating the reachable sets in discrete time (polygons with thick outlines) may still be insufficient for detecting collisions.
    To find an ego motion plan, we propose a numerical trajectory optimization approach with zonotope collision-avoidance constraints.
    The zonotopes are backed out from the outputs of a neural network predictive model, which also provides gradients for trajectory optimization (green dashed arrows).
    We also illustrate the ego goal $\state_\goal$, agents' state history $\statehistory$, and static obstacles $\statespace_\obs$.}
    \label{fig: front figure}
    \vspace*{-0.3cm}
\end{figure}

\textit{Related Work.}
A wide variety of methods predict future agent motion for navigation and interaction. 
To achieve high-quality predictions, it is typically necessary to explicitly model interactions, known dynamics, and multimodality \cite{gulzar2021survey,huang2022survey,karle2022scenario}. %
A further need across prediction architectures is to represent uncertainty in predicted states (e.g., \cite{koschi2020set,jacobs2017real,IvanovicElhafsiEtAl2020,Zhao_2019_CVPR,salzmann2020trajectron,suo2021trafficsim,girgis2021latent}).
For the sake of numerical tractability, it is common to either assume a Gaussian mixture model \cite{salzmann2020trajectron,IvanovicElhafsiEtAl2020}, or to associate occupancy probabilities with road network structure \cite{kaniarasu2020goal,koschi2020set}.
A key aspect of these methods is that they predict in \textit{discrete time} for numerical tractability (with the exception of \cite{koschi2020set}, which requires known road structure).

Given a prediction model, one can then perform planning and control.
In structured settings, one can adopt specific rules of the road as constraints for trajectory planning \cite{shalev2017formal,censi2019liability,undurti2011decentralized,vaskov2019not,vaskov2019towards,pek2020fail,sontges2017computing,johnson2012optimal}.
One can also consider \emph{risk tolerance} to allow a user-specified level of constraint violation \cite{undurti2011decentralized,nyberg2021risk}.
In less structured scenes, one can robustly consider interaction dynamics \cite{leung2020infusing,schaefer2021leveraging}; robustness can introduce conservativeness, which one may mitigate by estimating parameters of other agents' motion \cite{fridovich2020confidence}.
While most methods use discrete time, many planning methods also explicitly consider continuous time in unstructured \cite{vaskov2019towards} or structured \cite{vaskov2019not,althoff2014online,pek2020fail,sontges2017computing} settings.
However, these continuous-time methods typically only consider an \emph{open-loop} prediction model, wherein predictions may be incorrect because they are not conditioned on an ego trajectory plan.
Thus, the challenge we address is continuous-time, closed-loop collision avoidance in a numerically-tractable way.

\textit{Contributions.}
Our key insight is that typical discrete-time predictions can be readily extended to continuous-time via zonotopes, enabling a differentiable framework for optimization-based motion planning.
Our Zonotope Agreement of Prediction and Planning (ZAPP) is enabled by the following contributions.
First, we convert unbounded, multi-modal, discrete-time predictions into zonotopes suitable for robust planning.
Second, we extend discrete-time zonotope predictions to continuous time while enabling differentiable collision checking for trajectory optimization.
Third, we provide a simulation framework for constructing interactive scenes (see Fig. \ref{fig: zapp example}), and we show that ZAPP outperforms a variety of baselines through numerical experiments. %

\textit{Notation.}
The real numbers are $\R$ and the natural numbers are $\N$. 
The floor operator $\floor{\,\cdot\,}$ rounds down to the nearest integer.
The cardinality of $\set{A}$ is $\card{\set{A}}$.
The Minkowski sum is $\set{A} + \set{B} = \{a + b\ |\ a \in \set{A}, b \in \set{B} \}.$
A random variable $\noise$, drawn from an $n$-dimensional normal distribution with mean $\mu$ and covariance $\Sigma$, is $\noise \sim \normaldist(\mu,\Sigma)$.
The $n\times n$ identity matrix is $\eye_n$.
An $n\times m$ array of zeros (resp. ones) is $\zeros_{n\times m}$ (resp. $\ones_{n\times m}$).
For a vector $v$, the \ith element is $v\idx{i}$.
For an array $A$, the element at row $i$, column $j$ is $A\idx{i,j}$; the \ith row is $A\idx{i,:}$.
The $\diag$ operator places its arguments block-diagonally in a matrix of zeros.
\section{Problem Formulation}
\label{sec: problem statement}

We now pose a generic tightly-coupled prediction and planning problem (see Program \eqref{prog: general multiagent multimodal trajopt}).
This formulation requires solving a differential equation interaction model and collision checking in continuous time, necessitating a careful choice of representation for practical implementation.

\subsection{Setup}

\subsubsection{Agents, Modes, and Dynamics}
Consider a control scenario with an ego agent (which we control) and $\nagents \in \N$ other interacting agents (which we do not control, but we can predict their motion).
We assume an upper bound on $\nagents$ is known \emph{a priori}. %

We consider multi-modal interaction dynamics, where each mode $\mode \in \modes = \{1,2,\cdots,\nmodes\}$ is a unique homotopy class of the multi-agent system's trajectories, meaning that trajectories from one mode cannot be continuously defomed into those of another mode \cite{schmitzberger2002capture}.
For example, in a scene with two agents crossing paths (see Fig. \ref{fig: front figure}), there are at least two modes: one where agent $i$ stops to let the ego agent pass, and one where the ego agent stops to let agent $i$ pass.%
The ego agent has state $\state_\ego^\mode(t) \in \R^{n_\ego}$ in mode $\mode$ at time $t$.
The state vectors of the other $\nagents$ agents in mode $\mode$ are $\state_1^\mode(t),\cdots,\state_\nagents^\mode(t)$, with state dimensions $n_i \in \N$ for $i = 1,\cdots,\nagents$ respectively.
The combined state of the multi-agent system is $\state^\mode(t) \in \statespace \subset \R^\nstate$, with $\nstate = \nego + \sum n_i$.
We consider a finite time horizon $\horizon > 0$, so $t \in \timeinterval = [0,\horizon]$.

We assume that all agents occupy a 2-D workspace; our approach can extend to 3-D by applying techniques similar to \cite{kousik2019safe,holmes2020reachable,SelimAlanwarEtAl2022}.
We further assume that each agent's state $\state_i^\mode$ includes a 2-D \textit{position}, $\position_i^\mode(t) = \posproj_i(\state^\mode(t))$.
We denote static obstacles $\statespace_\obs \subset \statespace$, and assume they are known to the ego agent, since the focus of this work is planning.

We control the ego agent with a signal $\ctrl^\mode: \timeinterval \to \ctrlspace$. %
We associate the control signal with a particular mode to make this formulation general. %
We apply a receding-horizon strategy wherein $\ctrl$ is reoptimized every $\dt > 0$ seconds. %
We let $t$ be reset to $0$ every $\dt$ seconds at the beginning of each plan without loss of generality, so we only consider the time interval $\timeinterval = [0,\horizon]$ going forward.

Finally, we denote the multi-agent dynamics in mode $\mode$ as
\begin{align}\label{eq: cont time dyn}
    \dot{\state}^\mode(t) = \dyn^\mode(t,\state^\mode(t),\ctrl^\mode(t)) + \noise^\mode(t).
\end{align}
We assume the noise $\noise^\mode(t) \sim \noiseset$ is drawn from a distribution with compact support. %

\subsubsection{Reachable Sets}
We seek to find safe ego motion plans given the uncertain dynamics \eqref{eq: cont time dyn}.
To do so, we consider the reachable set of the multiagent system:
\begin{align}\label{eq: reachable set definition}
    \reachset^\mode(t) = \big\{ \state^\mode(t)\ |\ \dot{\state}^\mode = \dyn^\mode + \noise^\mode\ 
            \regtext{and}\ 
            \state^\mode(0) \in \reachset(0)
        \big\},
\end{align}
where $\reachset(0)$ is an uncertain set of initial states of all agents, which we assume is known.
Note, our proposed zonotope framework in Sec. \ref{sec: proposed method} can be readily extended to account for occupancy (nonzero volume) of each agent \cite{kousik2019safe,holmes2020reachable,althoff2014online,SelimAlanwarEtAl2022}, so we omit occupancy to simplify exposition.

\subsection{Problem Statement}

We seek to approximate the following trajectory optimization problem in a numerically tractable way:
\begin{subequations}\label{prog: general multiagent multimodal trajopt}
\begin{align}
    \min_{\{\ctrl^\mode\}_{\mode \in \modes}}\hspace{0.6cm}
        &\costfunc(\mode,\state^\mode,\ctrl^\mode)\\
    \regtext{s.t.}\hspace{0.9cm}
        &\posproj_\ego(\reachset^\mode(t)) \cap \posproj_i(\reachset^\mode(t)) = \emptyset~\forall\ t, \mode, i \label{prog constraint: dynamic collision avoidance} \\
        &\posproj_\ego(\reachset^\mode(t)) \cap \statespace_\obs = \emptyset~\forall\ t, \mode\label{prog constraint: static collision avoidance} \\
        &\ctrl^a(\tau) = \ctrl^b(\tau)\ \forall\ \tau \in [0,\consensushorizon], a, b \in \modes\label{prog constraint: consensus horizon}
\end{align}
\end{subequations}
where $\costfunc$ is an arbitrary cost function, $t \in \timeinterval$, $\mode \in \modes$, and $i \in \{1,\cdots,\nagents\}$.
Program \eqref{prog: general multiagent multimodal trajopt} seeks a set of $\card{\modes}$ control signals $\{\ctrl^\mode: \timeinterval \to \ctrlspace\}_{\mode \in \modes}$ for which the ego agent does not collide with any other agent (Constraint \eqref{prog constraint: dynamic collision avoidance}) or any obstacles (Constraint \eqref{prog constraint: static collision avoidance}).
Per \eqref{eq: reachable set definition}, the reachable set constraints \eqref{prog constraint: dynamic collision avoidance} and \eqref{prog constraint: static collision avoidance} implicitly require the multi-agent system to start from the initial condition set $\reachset(0)$ and obey the interaction dynamics $\dyn$.
Finally, the control signals must agree up to a \textit{consensus horizon} $\consensushorizon \in [\dt,\horizon]$ (Constraint \eqref{prog constraint: consensus horizon}); this means that the ego agent must apply the same control at least up to time $\consensushorizon$ for all modes.
We adopt this strategy because we do not necessarily know the actual mode when solving \eqref{prog: general multiagent multimodal trajopt}, so we must be able to simultaneously plan a control strategy for each mode.
This idea, Contingency MPC \cite{alsterda2019contingency}, has been applied successfully in multi-agent settings \cite{IvanovicElhafsiEtAl2020}.

The key challenge is that one can never perfectly represent the reachable sets $\reachset_i^\mode(t)$.
Furthermore, to enable a numerical solution, we require an efficient, differentiable representation of collision detection per \eqref{prog constraint: dynamic collision avoidance} and \eqref{prog constraint: static collision avoidance}.
\section{Proposed Method}
\label{sec: proposed method}

We now propose ZAPP to represent Problem \eqref{prog: general multiagent multimodal trajopt} in a numerically-tractable way.
We begin by introducing zonotopes and operations that we use to implement Problem \eqref{prog: general multiagent multimodal trajopt}
Then, we reformulate Problem \eqref{prog: general multiagent multimodal trajopt} as Problem \eqref{prog: trajopt numerical implementation} and detail our implementation.

\subsection{Zonotope Preliminaries}\label{subsec: zonotope prelims}
We use zonotopes to represent reachable sets.
A zonotope $\set{Z} \subset \R^n$ is a convex, symmetrical polytope parameterized by a \textit{center} $c \in \R^n$ and a \textit{generator matrix} $G \subset \R^{n\times \ngen}$ as
\begin{align}\label{eq: zono defn}
    \set{Z} = \zonofn{c,G} = \left\{c + G\beta\ |\ \norm{\beta}_\infty \leq 1 \right\},
\end{align}
where $\beta \in \R^\ngen$, and $\ngen$ is the number of \textit{generators} (i.e., the columns of $G$).
This is called the \textit{G-representation}.

Zonotopes enable efficient implementation of many set operations via well-known analytical formulas (see \cite{althoff2010reachability,guibas2003zonotopes}).
We use the Minkowski sum to construct continuous-time reachable sets and collision avoidance constraints.
The Minkowski sum of $\set{Z}_1 = \zonofn{c_1,G_1}$ and $\set{Z}_2 = \zonofn{c_2,G_2}$ is
$\set{Z}_1 + \set{Z}_2 = \zonofn{c_1 + c_2, [G_1, G_2]}$.
We use the Cartesian product to construct multi-agent reachable sets.
The Cartesian product is
$
    \set{Z}_1 \times \set{Z}_2 = \zonofn{
            [\begin{smallmatrix}
                c_1 \\ c_2
            \end{smallmatrix}],
        \diag{G_1, G_2}}.
$ %

We project zonotopes to lower dimensions to extract position information.
Suppose the multi-agent system is at state $\state(t) \in \zonofn{c(t),G(t)} \subset \R^\nstate$, where both $c(t)$ and $G(t)$ have $\nstate$ rows. %
We write $\position_i(t) \in \posproj_i(\zonofn{c(t),G(t)})$, which selects the rows of $c(t)$ and $G(t)$ corresponding to the position coordinates of state $\state_i$.

To detect if zonotope reachable sets are in collision,
one can check if the center of one zonotope lies in the other zonotope Minkowski summed with the generators of the first:
\begin{prop}[{\cite[Lemma 5.1]{guibas2003zonotopes}}]\label{prop: zono collision check}
Consider the zonotopes $\zonotope_1 = \zonofn{c_1,G_2}$ and $\zonotope_2 = \zonofn{c_2,G_2}$.
Then,
    $\zonotope_1 \cap \zonotope_2 = \emptyset$ if and only if
    $c_1 \not\in \zonofn{c_2,[G_1,G_2]}$.
\end{prop}
\noindent Note, checking if a point lies within a zonotope of arbitrary dimension typically requires either an iterative approach \cite{guibas2003zonotopes} or solving a linear program \cite{scott_constrained_2016,kulmburg2021co}.
In this work, we avoid this issue by leveraging two facts: (i) our occupancy sets are in 2-D, and (ii) since zonotopes are polytopes, they can also be represented as a collection of linear inequalities, also know as an H-representation, and represented by a matrix $A$ and a vector $b$ such that $x \in \set{Z} \iff \max(Ax - b) \leq 0$:

\begin{prop}[{\cite[Theorem 2.1]{althoff2010reachability}}]\label{prop: zono H-rep conversion}
Let $\set{Z} = \zonofn{c,G} \subset \R^2$, with $\ngen$ generators.
Assume that $G$ has no generators of length $0$.
Let $\ell_G \in \R^m$ be a vector of the lengths of each generator: $\ell_G\idx{i} = \norm{G\idx{:,i}}_2$.
Let $C = \left[\begin{smallmatrix*}
        -G\idx{2,:} \\ G\idx{1,:}
    \end{smallmatrix*}\right]$.
Then %
\begin{subequations}\label{eq: zono H-rep conversion A and b}
\begin{align}
    A\idx{:,i} &= \frac{1}{\ell_G\idx{i}} \cdot 
        \begin{bmatrix}
            C \\
            -C
        \end{bmatrix} \in \R^{(2\ngen)\times 2},
        \quad\regtext{and}\\
    b &= C\trans c +
    \abs{C\trans G}\ones_{m\times 1} \in \R^{2\ngen},
\end{align}
\end{subequations}
where $\abs{\cdot}$ denotes the absolute value taken elementwise.
Then
\begin{align}\label{eq: not in zono H-rep condition}
    x \notin \set{Z} \iff \max\big(Ax - b) > 0.
\end{align}
\end{prop}

\subsection{Reformulation for Implementation}

Let $\horizondisc = \floor{\horizon/\dt}$ and $\consensushorizondisc = \floor{\consensushorizon/\dt}$.
Suppose each mode $\mode$ is associated with a probability $\modeprobability^\mode \in [0,1]$ of occurring; in practice, we use a Gaussian mixture model (GMM) to learn multi-agent uncertain dynamics, so $\modeprobability^\mode$ are the coefficients of the GMM.
We implement the following program:
\begin{subequations}\label{prog: trajopt numerical implementation}
\begin{align}
    \min_{\{\variation\ctrlfuture^\mode\}_{\mode \in \modes}}
        &\sum_\mode
            \modeprobability^\mode \cdot \left(
                \lambda\lbl{f}\norm{\state^\mode(\horizondisc) - \state_\goal}_2^2 + 
                \sum_{k=0}^\horizondisc\lambda\lbl{r}\norm{\ctrl^\mode(k)}_2^2
            \right)\\
    \regtext{s.t.}\hspace{0.3cm}
        &\max\big(A_i^\mode(k)\position_\ego^\mode(k) - b_i^\mode(k)\big) > 0\ \forall\ i, k, \mode \label{prog constraint: dynamic collision avoidance implementation} \\
        &\max\big(A_j^\mode(k)\position_\ego^\mode(k) - b_j^\mode(k)\big) > 0\ \forall\ j, k, \mode \label{prog constraint: static collision avoidance implementation} \\
        &\ctrl^a(k) = \ctrl^b(k)\ \forall\ a, b \in \modes,\ k = 0,\cdots,\consensushorizondisc\label{prog constraint: consensus horizon implementation} \\
        &\variation\ctrl^\mode(k) + \ctrl_\nom^\mode(k) = \ctrl^\mode(k)\ \forall\ k, \mode \label{prog constraint: control perturbation definition}
\end{align}
\end{subequations}
where $i \in \{1,\cdots,\nagents\}$, $k \in \{0,\cdots,\horizondisc\}$, $\mode \in \modes$, and $\consensushorizondisc = \consensushorizon\dt$.
We set $x_\goal \in \statespace$ as a user-specified goal state and $\lambda\lbl{f},\lambda\lbl{r} > 0$ as user-specified weights.
The optimization variables $\variation \ctrlfuture^\mode \subset \varctrlspace \subset \R^\nctrl$ are perturbations of the control signal $\ctrl^\mode: \timeinterval \to \ctrlspace$ with respect to a fixed nominal control signal $u_\nom^\mode: \timeinterval \to \ctrlspace$, as in \eqref{prog constraint: control perturbation definition}; this decision variable formulation is detailed below in Sec. \ref{subsec: cost and decvar implementation}.
The collision avoidance constraints for dynamic obstacles \eqref{prog constraint: dynamic collision avoidance implementation} and static obstacles \eqref{prog constraint: static collision avoidance implementation} both leverage an H-representation, where the $A_\star^\star(k)$ and $b_\star^\star(k)$ matrices are constructed using Prop. \ref{prop: zono H-rep conversion} applied to zonotopes constructed in Sec. \ref{subsec: collision avoidance implementation}.
Static obstacles in \eqref{prog constraint: static collision avoidance implementation} are indexed by $j = 1,\cdots,n_\obs$.
To implement Program \eqref{prog: trajopt numerical implementation}, we use a gradient-based solver.
In the following, we are careful to ensure that \eqref{prog constraint: dynamic collision avoidance implementation}--\eqref{prog constraint: consensus horizon implementation} are differentiable with respect to $\{\variation\ctrlfuture^\mode\}_{\mode \in \modes}$.

\begin{rem}
We make a minor abuse of notation to write $k$ instead of $k\dt$ as arguments to time-varying quantities. %
\end{rem}

To proceed, first, we detail our decision variable implementation.
Then, we approximate discrete-time reachable sets using a prediction model.
Next, we extend from discrete to continuous time.
Finally, we discuss how to collision check our reachable sets.

\subsection{Dynamics via a Prediction Model}\label{subsec: prediction model}

\subsubsection{Reformulation}
It is often numerically intractable to perfectly represent the dynamics $\dyn^\mode$, and thus the reachable sets $\reachset^\mode(t)$, especially in arbitrary multi-agent settings.
Instead, we train a predictor model $\pred$ to estimate discretized solutions to \eqref{eq: cont time dyn} by maximizing the likelihood that
$
    \state^\mode(k) \sim \pred(\mode,k,\statehistory,\ctrlfuture^\mode;\param),
$
where $\param \in \R^{n_\param}$ are the trained parameters (e.g., neural network weights).
The sequence $\statehistory = \big(\state(k)\big)_{k=-\nhistory}^{0}$ is a finite, discrete state history of length $\nhistory \in \N$. %
Note that the states $\state(k)$ in the state history are not associated with a specific mode, since the past mode may not be known and the mode may change in the future.
Finally, $\ctrlfuture^\mode \subset \ctrlspace$ is the planned (future) control signal $\ctrl^\mode: \timeinterval \to \ctrlspace$.

\subsubsection{Implementation}
We use Trajectron++ \cite{salzmann2020trajectron}, a state-of-the-art prediction model implemented as a Conditional Variational Auto-Encoder (CVAE).
Trajectron++ assumes a known form of dynamics and control for each agent (e.g., 2-D integrator or kinematic unicycle). %
The model outputs a multi-modal distribution over each agent's applied controls at each time step, represented as a Gaussian mixture, with one multi-variate Gaussian per mode: $\ctrl_i^\mode(k) \sim \normaldist(\mean_{i,\ctrl}^\mode(k), \std_{i,\ctrl}^\mode(k))$.
Here, $\mean_{i,\ctrl}^\mode(k)$ is the mean over controls for agent $i$ in mode $\mode$, and $\std_{i,\ctrl}^\mode(k)$ is the covariance matrix.
Trajectron++ produces a distribution over each agent's state by propagating the control and previous state distributions forward according to each agent's dynamics, which we denote by the mappings:
\begin{subequations}
\begin{align}
\label{eq: trajectron mapping}
    g_\mean:\ 
        (\mode,k,\statehistory,\ctrlfuture^\mode) &\mapsto \mean_i^\mode(k)\ \regtext{and}\\
    g_\std:\ 
        (\mode,k,\statehistory,\ctrlfuture^\mode) &\mapsto \std_i^\mode(k),
\end{align}
\end{subequations}
where $\mean_i^\mode(k) \in \R^{n_i}$ is a mean state and $\std_i^\mode(k) \in \R^{n_i\times n_i}$ is a state covariance matrix, such that
$
    \state_i^\mode(k) \sim \normaldist(\mean_i^\mode(k),\std_i^\mode(k))
$
with high likelihood.
Note that $\mean_i^\mode$ (mean state) is denoted differently from $\mean_{i,\ctrl}^\mode$ (mean control), and similarly for standard deviation.

\subsection{Decision Variable Implementation}\label{subsec: cost and decvar implementation}

When solving Problem \eqref{prog: trajopt numerical implementation} numerically, our predictor would need to be re-evaluated after each solver iteration, because the output prediction is conditioned on the future ego motion.
In other words, one must perform a forward-pass of a neural network multiple times for a single MPC solve, which can be computationally burdensome \cite{schaefer2021leveraging}.
To avoid this issue, we implement our decision variables as follows.

First, the predictor is evaluated once at the beginning of an MPC execution using $U_{\regtext{f}, \nom}^\mode \subset \ctrlspace$, which is a fixed nominal control $u_\nom^\mode: \timeinterval \to \ctrlspace$.
In practice, we create this nominal control as an open-loop sequence of controls that drives the ego agent in a straight line towards its global goal (ignoring static or dynamic obstacles).
We obtain nominal mean states $\mean_{i, \nom}^\mode$ and nominal state covariance matrices $\std_{i, \nom}^\mode$ as
\begin{subequations}
\begin{align}
    \mean_{i, \nom}^\mode(k) &= g_\mean(\mode,k,\statehistory,U_{\regtext{f}, \regtext{nom}}^\mode),\\
    \std_{i, \nom}^\mode(k) &= g_\std(\mode,k,\statehistory,U_{\regtext{f}, \regtext{nom}}^\mode),
\end{align}
\end{subequations}
which are created by rolling out the Trajectron++ predictor dynamics \cite{salzmann2020trajectron}.
Then, we consider a first-order Taylor expansion around the nominal control, so our decision variables are $\variation \ctrlfuture^\mode$ such that
\begin{align}\label{eq: taylor expanded means and covariances}
    \expansion{\mean}_i^\mode(k) &\coloneqq 
    \mean_{i, \nom}^\mode(k) + \left.\tfrac{\partial g_\mean}{\partial \ctrlfuture^\mode}\right|_{(k, \statehistory, U_{\regtext{f}, \regtext{nom}}^\mode)}\ \variation \ctrlfuture^\mode,
\end{align}
where $\expansion{\mean}_i^\mode(k)$ is the new perturbed mean state.

Optimizing and constraining the perturbations of the control sequence, instead of the control sequence itself, ensures that the updated solutions provided by the solver remain fairly close to the nominal control sequence, which is used for the evaluation of the neural network.
Since we use a receding-horizon MPC framework, we take advantage of warmstarting each time we call our solver (i.e., to initialize our decision variable, we take the optimal control sequence from the previous solution, discard the value of the first timestep, and duplicate the value of the last timestep).

\subsection{Discrete-Time Reachable Sets via a Prediction Model}\label{subsec: discrete time reachability}

\subsubsection{Reformulation}
We seek to approximate each discrete-time reachable set $\reachset^\mode(k)$ as an $\alpha$-confidence region of the predictor.
We extend this to continuous time in Sec. \ref{subsec: cont time approx}.

\subsubsection{Implementation}
Given the distribution from Trajectron++, we represent the reachable set as a zonotope overapproximating a confidence region of the distribution:
\begin{align}\label{eq: network output zonotope}
    \reachset_i^\mode(k) \approx \zonotope^\mode_i(k) = \zonofn{\expansion{\mean}^\mode(k), \expansion{G}_\std^\mode(k)},
\end{align}
with
$
    \expansion{\mean}^\mode(k) = 
        (\expansion{\mean}_\ego^\mode(k),\expansion{\mean}_i^\mode(k),\cdots,\expansion{\mean}_\nagents^\mode(k))
$, and
$
    \expansion{G}_\std^\mode(k) =
        \diag(\expansion{G}_\ego^\mode(k),\expansion{G}_1^\mode(k),\cdots,\expansion{G}_\nagents^\mode(k)).
$
In particular, we construct the generators in each $\expansion{G}_i^\mode(k)$ as the principal axes of a confidence ellipsoid of the distribution $\normaldist(\expansion{\mean}_i^\mode(k),\expansion{\std}_i^\mode(k))$, per \cite{shetty2020trajectory} and \cite[Proposition 2]{althoff2009safety}.
We set
$\expansion{G}_i^\mode(k) = \vep_i\left[(\lambda_{i,1}^\mode)^{1/2}v_{i,1}^\mode,\cdots,(\lambda_{i,n_i}^\mode)^{1/2}v_{i,n_i}^\mode \right]$ and
$\vep_i = ({\regtext{chi2inv}}_{n_i}\!\left(\regtext{erf}(\alpha/\sqrt{2})\right))^{1/2}$, where $\lambda_{i,j}^\mode$ is the \jth eigenvalue, $v_{i,j}^\mode$ is the \jth eigenvector for agent $i$ in mode $\mode$, $\regtext{chi2inv}_{n_i}$ is the inverse of the $\chi^2$ (chi squared) distribution function with $n_i$ degrees of freedom, and $\regtext{erf}$ is the error function \cite[Sec. 3.1]{wang2015confidence}.
Note that $\vep_i$ can be precomputed for a user-specified confidence bound for each agent, and is constant when solving Problem \eqref{prog: trajopt numerical implementation}.
We set $\alpha = 1.0$.

\subsection{Continuous-time Reachable Set Approximation}\label{subsec: cont time approx}

We now approximate the continuous-time reachable sets from \eqref{prog constraint: dynamic collision avoidance} and \eqref{prog constraint: static collision avoidance} using zonotopes, which we use later in Sec. \ref{subsec: collision avoidance implementation} to generate the constraints \eqref{prog constraint: dynamic collision avoidance implementation} and \eqref{prog constraint: static collision avoidance implementation}.

\subsubsection{Reformulation}
We seek to model the continuous-time reachable set for each agent, denoted
\begin{align}\label{eq: cont reach set defn}
    \contreachset_i^\mode(k) =
        \union_{\tau \in \left[k\dt,\, (k+1)\dt\right]} \reachset_i^\mode(\state^\mode(\tau)).
\end{align}

\subsubsection{Implementation}
We propose a geometric approximation
of \eqref{eq: cont reach set defn}.
We construct $\cont{c}_i^\mode(k)$ and $\cont{G}_i^\mode(k)$ such that
\begin{align}\label{eq: cont reach set zono approx}
    \contreachset_i^\mode(k) &\approx
        \contzono_i^\mode(k) =
        \zonofn{\cont{c}_i^\mode(k),\cont{G}_i^\mode(k)}
\end{align}
The joint continuous-time reachable set for all agents is then
\begin{align}\label{eq: cont multiagent reach set zono approx}
    \contreachset^\mode(k) \approx
        \contzono^\mode(k) = 
        \contzono_\ego^\mode(k)\times \contzono_1^\mode(k) \times \cdots \times \contzono_\nagents^\mode(k).
\end{align}

To approximate $\contreachset_i^\mode(k)$ geometrically, we first represent the line segment between $\state_i^\mode(k)$ and $\state_i^\mode(k+1)$ as a zonotope:
\begin{align}\label{eq: line segment between states}
    \linesegment_i^\mode(k) &= \zonofn{c_{\linesegment,i}^\mode(k), G_{\linesegment,i}^\mode(k)},\ \regtext{with} \\
    c_{\linesegment,i}^\mode(k) &=
        \tfrac{1}{2}\big(\mean_i^\mode(k) + \mean_i(k+1)\big)\ \regtext{and} \\
    G_{\linesegment,i}^\mode(k) &=
        \tfrac{1}{2}\big(\mean_i^\mode(k+1) - \mean_i^\mode(k)\big).
\end{align}
Then, we approximate $\contreachset_i^\mode$ by extending the reachable set at each time step halfway towards the reachable sets at the previous and subsequent time steps:
\begin{align}\begin{split}\label{eq: cont occupancy approximation formulation}
    \contzono_i^\mode(k) =\  
        &\big(\zonotope_i^\mode(k) + \tfrac{1}{2}\big(\linesegment_i^\mode(k) - \state_i^\mode(k)\big)\big)\ \cup \\
        &\big(\zonotope_i^\mode(k+1) + \tfrac{1}{2}\big(\linesegment_i^\mode(k) - \state_i^\mode(k+1)\big)\big),
\end{split}\end{align}
where $\zonotope_i^\mode(k) = \zonofn{\hat{\mean}_i^\mode(k), \hat{G}_i^\mode{k}}$, and we have used the Minkowski sum of zonotopes as per Sec.~\ref{subsec: zonotope prelims}. %

We leave alternative methods of approximating continuous time to future work.
For example, one could compute the convex hull between timesteps with constrained zonotopes \cite{scott_constrained_2016,raghuraman_set_2020}, but the resulting numerical representations are typically large.
One could also apply standard zonotope reachability methods \cite{althoff2010reachability,althoff2014online}, though the resulting set representations may not be differentiable.

\subsection{Collision Avoidance Constraints}\label{subsec: collision avoidance implementation}

We now use our reachable set approximation to create the collision avoidance constraints \eqref{prog constraint: dynamic collision avoidance implementation} and \eqref{prog constraint: static collision avoidance implementation}.

\subsubsection{Reformulation}
We reformulate the collision avoidance constraints as
\begin{subequations}\label{eq: collision reformulation}
\begin{align}
    &\posproj_\ego(\contreachset^\mode(k)) \cap \posproj_i(\contreachset^\mode(k)) = \emptyset 
        \quad\regtext{and}\quad \\
    &\posproj_\ego(\contreachset^\mode(k)) \cap \statespace_\obs = \emptyset
\end{align}
\end{subequations}
for each $k = 0,\cdots,\horizondisc-1$.

\subsubsection{Implementation}

We construct the dynamic collision avoidance constraints as follows.
Consider the ego agent and agent $i$ in mode $\mode$ at time step $k$.
Recall that the continuous time occupancy approximation is given as $\contzono_i^\mode(k) = \zonofnplain(\cont{c}_i^\mode(k),\cont{G}_i^\mode(k))$ for agent $i$.
We apply Prop. \ref{prop: zono collision check} and project each such zonotope to the agent's position dimensions to construct a collision check zonotope between the ego and each \ith agent:
\begin{align}
    \poszono_{i}^\mode(k) = 
        \zonofnplain\big(\position_i^\mode(k),
                \big[\posproj_\ego(\cont{G}_\ego^\mode(k)),
                    \posproj_i(\cont{G}_i^\mode(k))\big]
            \big),
\end{align}
where $\posproj_i$ extracts the position coordinates of the generator matrix.
We convert $\poszono_{i}^\mode(k)$ to an H-representation $(A_i^\mode(k), b_i^\mode(k))$ with Prop. \ref{prop: zono H-rep conversion} to get the constraint in \eqref{prog constraint: dynamic collision avoidance implementation}:
\begin{align}
    \max\big(A_i^\mode(k)\position_\ego^\mode(k) - b_i^\mode(k)\big) > 0.
\end{align}

We formulate the static obstacle collision avoidance constraints similarly.
First, we assume the static obstacles are overapproximated by a finite union of zonotopes: $\statespace_\obs \subseteq \union_{j=1}^{n_\obs} \zonofn{\position_{\obs,j}, G_{\obs,j}}$.
This is a reasonable assumption for a variety of common obstacle representations, such as occupancy grids or compact polygons.
Then, for each obstacle $j$, we apply Prop. \ref{prop: zono collision check} to construct a zonotope
    $\poszono_{j}^\mode(k) = 
        \zonofnplain\big(\position_{\obs,j},
                \big[\posproj_\ego(\cont{G}_\ego^\mode(k)), G_{\obs,j}\big]
            \big)$.
We apply Prop. \ref{prop: zono H-rep conversion} to convert to H-representation, $(A_j^\mode(k), b_j^\mode(k))$, as in \eqref{prog constraint: static collision avoidance implementation}.

Importantly, our collision avoidance constraints are subdifferentiable \cite[Ch. 5.1.4]{polak2012optimization}, meaning we can use them with a gradient-based optimization solver.
This is because we leverage Props. \ref{prop: zono collision check} and \ref{prop: zono H-rep conversion}; notice that Prop. \ref{prop: zono collision check} only requires matrix concatenation, and Prop. \ref{prop: zono H-rep conversion} uses arithmetic, concatenations, an absolute value, and a max.
In Prop. \ref{prop: zono H-rep conversion}, generator lengths appear in the denominator of \eqref{eq: zono H-rep conversion A and b}, but we assume there are no generators of length 0.
In practice, we compute derivatives of the constraints with respect to the ego controls using autodifferentiation in PyTorch \cite{pytorch}.

Next, we present a numerical experiment to illustrate the utility of our proposed ZAPP approach.
\section{Numerical Experiments}\label{sec: numerical examples}

We now present a numerical study to illustrate the utility of our proposed method for uncertainty-aware planning in multi-agent scenes.
As part of our codebase, we contribute a framework for generating interactive scenes.
We created this because we found that, in preliminary experiments with traffic data~\cite{nuscenes,highway_env}, traffic agents mostly follow predefined paths with little interaction.
Our code is open-source\footnote{\url{https://github.com/lpaparusso/ZAPP}}.
We implement zonotopes in python using the numpy package \cite{harris2020array}.
We solve \eqref{prog: trajopt numerical implementation} using the Interior Point Method (IPM) solver cyipopt\footnote{
\url{https://cyipopt.readthedocs.io/en/stable/}
}.
The cost and constraints gradients are computed via PyTorch \cite{pytorch} automatic differentiation.

We generate our dataset using the environment shown in Figs. \ref{fig: zapp example} and \ref{fig: zapp full scene}. %
The goal is to navigate a hallway while avoiding 10 non-ego agents.
All agents are modeled as double integrators subject to nonlinear forces (note that Trajectron++~\cite{salzmann2020trajectron} can readily handle more complex nonlinear dynamics; we deliberately chose dynamics that allowed us to construct highly interactive scenes).
Each  $(i,j)$ pair of non-ego agents repels each other proportional to the inverse squared Euclidean distance $1/\norm{\position_i^\mode - \position_j^\mode}_2^2$.
They are also repelled, but with much lower magnitude, from the ego agent; this mimics social forces while allowing collisions (see Table \ref{tab: results}).
Finally, non-ego agents are repelled from static obstacles proportional to their relative velocity and inverse of relative distance.
This setup renders the prediction problem control-dependent and prevents the surrounding agents from easily avoiding the ego agent, ensuring interaction.
All agents have a maximum absolute velocity of 4 m/s and a maximum absolute acceleration of 3 m/s$^2$ in both the Cartesian directions.
We consider the ego agent to have no size (i.e., a point mass), and the surrounding agents to be squares with a side length of 1 m.

We train the predictor on 300 scenes of 8 seconds each. 
In the training scenes, the ego controls are simulated by creating an artificial potential field that attracts the ego agent towards the global goal, resulting in realistic and smooth control policies.

\subsection{Experiment Setup}
\subsubsection{Task}
We consider the control task complete when the ego agent has traveled 28 m horizontally from the start point.
If a collision involving the ego agent occurs, we register a crash and consider the task not complete.
Note, we collision check the robot approximately in continuous time by checking at a much finer time discretization than is used for our MPC planner.

\subsubsection{Planner Details}
Our planner considers the 2 most probable modes for each of the 3 closest surrounding agents.
It runs at 2 Hz in simulation time, with the 16 step predictive horizon discretized at 10 Hz.
We set $|\variation\ctrl^\mode(k)| \in [-3, 3]$ m/s$^2$ to ensure that the linearization of the prediction model is a fair assumption (see \eqref{prog constraint: control perturbation definition}).
We set the maximum number of solver iterations per MPC execution to 10.

\subsubsection{Baselines}
(i) To understand the relevance of continuous-time approximation, we compare against MATS, a planning-focused predictor that uses linearized dynamics and discrete-time collision avoidance \cite{IvanovicElhafsiEtAl2020}; to ensure a fair comparison, we augment MATS with our discrete-time reachable set approximation.
(ii) To test the effect of interaction prediction, we test against ZAPP without interaction gradients (w/o Int.), meaning we do not update the predictions when the ego control changes while solving Problem \eqref{prog: trajopt numerical implementation}. %
(iii) To test robustness to model error, we evaluate both MATS and ZAPP without domain consistency (w/o DC), meaning the predictor is trained on different interaction forces from the ones used in the experiment.
Future work will compare against additional social navigation baselines \cite{leung2020infusing,fisac2018probabilistically}.

\subsection{Results and Discussion}
\begin{figure}[t]
    \centering
    \includegraphics[width=0.90\columnwidth]{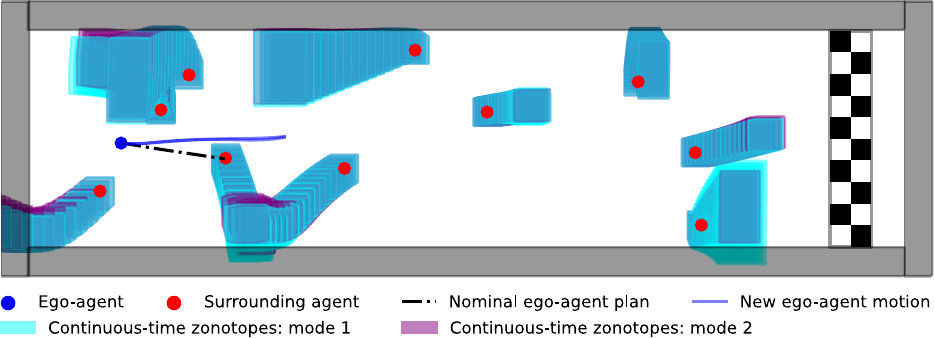}
    \caption{\small
    ZAPP takes as input a nominal ego-agent trajectory (black dashed line), which is used to predict the nominal continuous-time predictions (blue tubes), and finally steers the motion plan to avoid collision (blue line).}
    \label{fig: zapp example}
    \vspace*{-0.2cm}
\end{figure}

\begin{figure}[t]
    \centering
    \includegraphics[width=0.90\columnwidth]{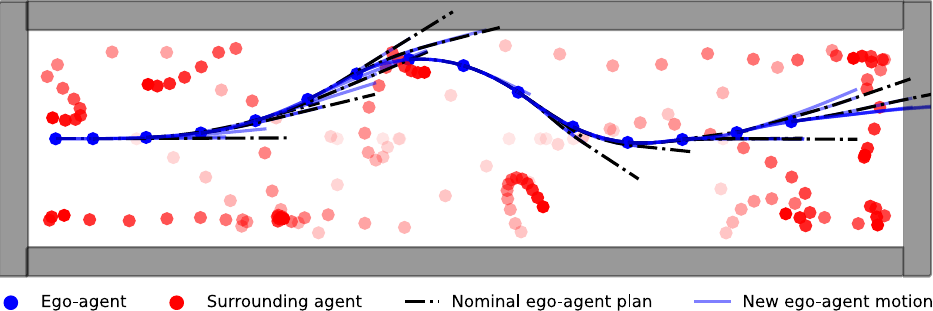}
    \caption{\small
    Example of a completed scenario in the robustness test (w/o DC).
    ZAPP is able to lead ego to the goal (from left to right) without crashes, implementing receding horizon motion planning (trajectories at each time step shown in blue).
    To ease visualization, the surrounding agents' trajectories are shown at discrete times with color fading from dark to light as time passes.}
    \label{fig: zapp full scene}
    \vspace*{-0.2cm}
\end{figure}

Quantitative results are shown in Table \ref{tab: results}. 
In Figure \ref{fig: zapp example}, a qualitative example shows ZAPP in action. A full completed scenario is shown in Figure \ref{fig: zapp full scene} for the ZAPP w/o DC case.

\paragraph{Crash Percentage}
ZAPP produces a low crash percentage (approximately 1/10 compared to MATS).
This shows that collision checking in continuous time has a positive effect on safety and task completion.
The fact that ZAPP w/o DC has more crashes than ZAPP indicates that modeling uncertainty (i.e., reachable sets) is necessary, and that improving predictor robustness is critical for future work. %
This is further supported by the high crash percentage of \text{MATS w/o DC}, which is not acceptable in safety-critical control tasks. 
Overall, these results indicate that the extension of collision checking to continuous time has a positive effect on safety and robustness to perturbations when planning ego motion in dynamic, multi-agent scenes.

\paragraph{Average Speed}
Table \ref{tab: results} reports the average speed (towards the goal) of the ego agent over all scenes with no crashes, similar to the metric in \cite{vaskov2019towards}.
The 4 baselines have similar average speed values, close to the maximum allowed control action of 4 m/s.
Therefore, ZAPP improves safety and robustness compared to MATS at practically no expense in task completion speed.
However, we stress that the metrics are calculated only for scenarios without crashes.
Since ZAPP completes the task almost always, its average speed is more relevant than the one obtained by MATS. 

\paragraph{Solve Time}
MATS baselines, as expected, compute faster because discrete-time collision checking needs half as many zonotopes and constraints.
However, all of the obtained solver times are only one order of magnitude slower than real time, even though the predictor and planner are written in non-compiled and non-optimized Python code.
Thus, we anticipate that, in future work, our method can readily be made to work in real time with compiled and optimized code, and task-focused prediction architectures (e.g., \cite{kamenev2022predictionnet}).

\begin{table}[t]
    \centering
    \begin{tabular}{l|c|c|c}
         Method & G/C [\%] & AS [m/s] & ST [s]\\
         \hline
         MATS~\cite{IvanovicElhafsiEtAl2020}           & 70.0 / 30.0 & 3.81$\pm$0.10 & 2.27$\pm$0.62 \\
         MATS~\cite{IvanovicElhafsiEtAl2020} w/o DC        & 53.3 / 46.7 & 3.79$\pm$0.14 & 2.27$\pm$0.62 \\
         ZAPP w/o DC      & 86.7 / 13.3 & 3.75$\pm$0.15 & 4.33$\pm$1.72 \\ 
         ZAPP w/o Int.     & 46.7 / 53.3 & \textbf{3.86$\pm$0.08} & \textbf{2.21$\pm$0.58} \\
         ZAPP (ours)         & \textbf{96.7 / 3.3} & 3.80$\pm$0.14 & 4.33$\pm$1.72                 
    \end{tabular}
    \caption{Baseline performance on goals/crashes (G/C), average speed (AS), and average solve time (ST).}
    \label{tab: results}
    \vspace*{-0.3cm}
\end{table}
\section{Conclusion}

This paper presented a Zonotope Agreement of Prediction and Planning (ZAPP) to overcome the gap between prediction and planning numerical representations for mobile robot motion planning in interactive scenes.
The method leverages zonotopes to enable continuous-time reasoning for planning, whereas most prediction frameworks are based on discrete time.
Numerical experiments show that this enables increased safety, and a significant safety boost over discrete time planning, when generating interactive motion plans in mobile robot settings.
Future work will explore alternative methods of extending from discrete to continuous time and alternative representations of reachable sets and uncertainty.
We note that ZAPP can extend to include strict safety guarantees with techniques such as \cite{vaskov2019towards,vaskov2019not,liu2022refine,shao2021reachability,SelimAlanwarEtAl2022}.
Furthermore, we plan to conduct hardware trials to validate the proposed approach.

\renewcommand{\bibfont}{\normalfont\footnotesize}
{\renewcommand{\markboth}[2]{}%
\printbibliography}

\end{document}